\documentclass[final,5p,times,twocolumn]{elsarticle}

\usepackage{amssymb}
\usepackage{amsmath}

\usepackage{algorithm}
\usepackage{algorithmic}
\usepackage{float}
\usepackage{subfigure}
\usepackage{url}
\usepackage{geometry}

\journal{Expert Systems with Applications}

\begin{document}

\begin{frontmatter}



\title{Improving Aviation Safety Analysis: Automated HFACS Classification Using Reinforcement Learning with Group Relative Policy Optimization}


\author[1]{Arash Ahmadi\corref{cor1}}
\ead{arash.ahmadi-1@ou.edu}

\author[1]{Sarah S. Sharif}
\ead{s.sh@ou.edu}

\author[1]{Yaser M. Banad}
\ead{bana@ou.edu}

\cortext[cor1]{Corresponding author}

\affiliation[1]{organization={School of Electrical, and Computer Engineering, University of Oklahoma},
            addressline={}, 
            city={Norman},
            postcode={73019}, 
            state={OK},
            country={USA}}

\begin{abstract}
  Analyzing the human factors behind aviation accidents is crucial for preventing future incidents, yet traditional methods using the Human Factors Analysis and Classification System (HFACS) are limited by scalability and consistency. To address this, we introduce an automated HFACS classification framework for aviation safety analysis that utilizes Reinforcement Learning with Group Relative Policy Optimization (GRPO) to fine-tune a Llama-3.1 8B language model. Our approach incorporates a multi-component reward system tailored for aviation safety analysis and integrates synthetic data generation to overcome class imbalance in accident datasets. The resulting GRPO-optimized model achieved noticeable performance gains, including a 350\% increase in exact match accuracy (from 0.0400 to 0.1800) and an improved partial match accuracy of 0.8800. Significantly, our specialized model outperforms state-of-the-art LLMs (Large Language Models), including GPT-5-mini and Gemini-2.5-flash, on key metrics. This research also proposes exact match accuracy in multi-label HFACS classification problem as a new benchmarking methodology to evaluate the advanced reasoning capabilities of language models. Ultimately, our work validates that smaller, domain-optimized models can provide a computationally efficient and better solution for critical safety analysis. This approach makes powerful, low-latency deployment on resource-constrained edge devices feasible.
\end{abstract}





\end{frontmatter}




\section{Introduction}

General aviation (GA) provides indispensable capacity for disaster relief \cite{morales2015managing}, emergency response \cite{wu2016emergency}, fixed-wing aeromedical transport \cite{sherry2022statistical}, long-running environmental monitoring using light aircraft \cite{kuo2017search}, and remote regions \cite{e2022geographical}.

Given this expanding operational footprint, strengthening GA safety is central to the resilience of the broader civil aviation system. The heterogeneity of GA missions creates diverse operational contexts and a rapidly growing volume of safety data. Understanding where and why breakdowns occur therefore requires methods that integrate established human-factors taxonomies with scalable, and data-driven inference.

Aviation safety analysis stands at a critical juncture where traditional human factors frameworks can intersect with cutting-edge artificial intelligence optimization techniques. The Human Factors Analysis and Classification System (HFACS) \cite{shappell2000human} has emerged as the dominant methodology for systematic accident investigation. However, traditional HFACS implementation faces significant scalability and consistency challenges that limit its effectiveness in modern aviation environments, where the volume and complexity of safety data continue to grow exponentially. These limitations highlight the need for adaptive, data-driven approaches that extend beyond conventional taxonomy-driven methods and open the door to more advanced forms of computational safety reasoning.

The emergence of large language models (LLMs) \cite{naveed2025comprehensive} has demonstrated transformative potential for automating complex safety reasoning tasks. Recent research \cite{liu2025accident} introduces HFACS-guided chain-of-thought prompting \cite{wei2022chain} that integrates systematic human factors analysis with Chain-Of-Thought technique which achieves expert-level performance in general aviation accident investigation. Similarly, AccidentGPT \cite{wu2024accidentgpt} extends this paradigm to traffic accident analysis by leveraging multi-modal inputs and privacy-preserving design for automatic and objective safety investigation

Despite being remarkable in fluency, most LLMs remain constrained by their tendency to reproduce memorized patterns rather than invent new reasoning strategies. Reinforcement learning (RL) \cite{sutton1998reinforcement} has long served as a cornerstone methodology for training intelligent systems by optimizing decision-making through trial-and-error interactions with an environment. Early value based approaches such as Q-learning \cite{watkins1992q} established fundamental principles for action-value estimation, while subsequent methods like SARSA \cite{wang2013backward} and Deep Q-Networks (DQN) \cite{mnih2013playing} extended these ideas to more complex and high dimensional tasks like playing Atari games. Improvements such as Double DQN (DDQN) \cite{van2016deep, halat2024modified} helped mitigate overestimation bias in value functions, further enhancing performance in challenging environments. The integration of deep neural networks with RL, often referred to as deep reinforcement learning \cite{li2017deep}, enabled scalable solutions for domains with large state and action spaces that laid the foundation for modern advancements. Actor-critic methods, including Advantage Actor-Critic (A2C) \cite{babaeizadeh2016reinforcement}, introduced hybrid frameworks that combined the strengths of value-based and policy-gradient techniques. Policy-gradient algorithms such as Trust Region Policy Optimization (TRPO) \cite{hariharan2025reinforcement} and Proximal Policy Optimization (PPO) \cite{schulman2017proximal} further improved training stability and scalability which enabled their application to challenging domains like natural language processing.

The superiority of reinforcement learning over traditional supervised fine-tuning for complex reasoning tasks has been demonstrated through several groundbreaking applications. AlphaGo's historic victory over world champion Go players \cite{silver2017mastering} showcased how RL enables systems to discover novel strategies and creative solutions that transcend human knowledge, going beyond mere pattern recognition to develop innovative approaches through self-play and exploration. This creative potential of RL extends far beyond game playing, as evidenced by recent advances in mathematical reasoning where smaller models trained with reinforcement learning can outperform significantly larger models trained with conventional methods. The DeepSeek-Math series \cite{shao2024deepseekmath} demonstrates this paradigm shift, where a 7 billion parameter model enhanced with Group Relative Policy Optimization (GRPO) achieves performance comparable to or exceeding that of much larger models like OpenAI's o1 \cite{jaech2024openai}, which contains orders of magnitude more parameters. This remarkable efficiency gain shows a fundamental limitation of supervised learning approaches: while they excel at mimicking patterns in training data, they struggle to generate truly novel reasoning pathways or discover creative solutions to complex problems.

Group Relative Policy Optimization (GRPO) \cite{shao2024deepseekmath} offers significant computational advantages through its novel approach to reinforcement learning for language model training. GRPO eliminates the need for separate value function models, that substantially reducing memory and computational burden compared to traditional Proximal Policy Optimization (PPO) approaches while achieving strong performance on complex reasoning tasks. The method's group sampling approach with relative advantage estimation leverages the comparative nature of reward models by normalizing rewards within groups. This is crucial for tasks requiring multi-step reasoning where traditional supervised fine-tuning may plateau. GRPO demonstrates that reinforcement learning can enhance model performance even when starting from already strong instruction-tuned baselines.

\begin{figure*}[t]
\centering
\includegraphics[width=0.9\textwidth]{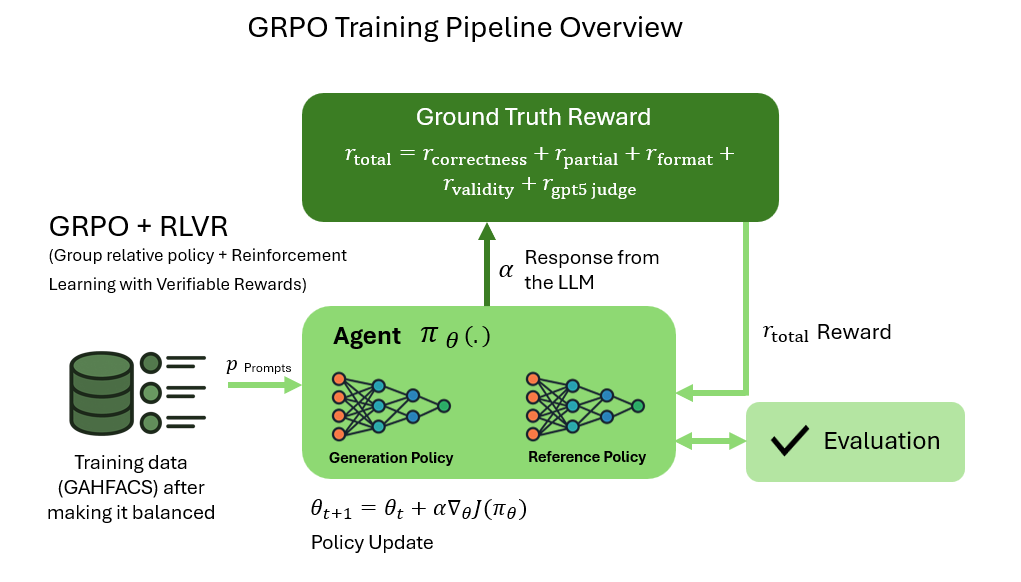}
\caption{GRPO Training Pipeline Overview for HFACS Classification. The system processes balanced GAHFACS training data through an agent with generation and reference policies. Multiple reward functions evaluate model responses including correctness, partial match, format compliance, validity, and GPT-5 reasoning quality assessment. The total reward feeds back into policy updates using group relative advantage estimation, while continuous evaluation monitors training progress.}
\label{fig:grpo_overview}
\end{figure*}

Figure \ref{fig:grpo_overview} illustrates our comprehensive training pipeline that integrates GRPO optimization with HFACS classification through a multi-component reward system. Our approach incorporates five distinct reward components: correctness rewards for exact HFACS code matches, partial match rewards, format rewards ensuring proper reasoning structure, validity rewards penalizing invalid and hallucinated codes, and gpt5-nano \cite{openai2025gpt5} judge rewards that evaluates reasoning quality. 

This work addresses the fundamental challenge of developing computationally efficient, and accurate language model for HFACS classification in aviation safety analysis. We propose a novel integration of GRPO with chain-of-thought reasoning that maintains the methodological rigor of traditional HFACS analysis while leveraging the efficiency advantages of modern reinforcement learning techniques.

The contributions of this research include: (1) the application of GRPO to aviation safety classification tasks, achieving significant computational efficiency gains over traditional approaches while demonstrating that a smaller language model trained with GRPO outperforms general-purpose large language models; (2) a comprehensive multi-component reward system with an LLM as a judge reward function that captures the nuanced requirements of HFACS classification; (3) integration of synthetic data generation techniques to address class imbalance challenges in aviation safety datasets; and (4) proposing exact match accuracy in HFACS classification as a new benchmarking methodology for evaluating LLM reasoning capabilities on complex multi-label classification tasks.

\section{Methodology}

This section presents our methodology for applying Group Relative Policy Optimization (GRPO) to HFACS classification in aviation safety analysis. We begin by describing the HFACS framework and our dataset construction approach, followed by a detailed explanation of GRPO and its adaptation for our classification task.

\subsection{HFACS Framework and Problem Formulation}

The Human Factors Analysis and Classification System (HFACS) provides a structured approach to categorizing human errors in aviation accidents . We focus on the first two layers of HFACS 8.0 \cite{liu2025accident}: unsafe acts and their preconditions. This subset was chosen because general aviation accidents often involve private flights and rarely include organizational factors that would require the full four-layer HFACS framework. The unsafe acts include Performance/Skill Based Errors (AE100), Judgment/Decision-making Errors (AE200), and Known Deviations (AD000). The preconditions encompass pilot conditions (PC100-PC300), environmental factors (PE100-PE200), planning conditions (PP100), and training conditions (PT100).

Our problem is formulated as a multi-label classification task where given an accident narrative $x$, the model must predict a set of relevant HFACS codes $y = \{y_1, y_2, ..., y_k\}$ where each $y_i$ belongs to the predefined HFACS categories. Our approach utilizes Llama 3.1 8B using Unsloth \cite{unsloth_2025_llama_8b_instrcut} as the base architecture, which we optimize using Group Relative Policy Optimization to improve classification accuracy and reasoning quality.

The multi-label formulation is essential for capturing the complex nature of aviation safety incidents. Aviation accidents rarely result from a single isolated human factor. Instead, they typically emerge from complex interactions between multiple failure modes across different HFACS categories. For example, a pilot's skill-based error (AE100) may be compounded by inadequate training conditions (PT100) and adverse environmental factors (PE100), which creates a chain of events that leads to an accident. A single-label approach would force the model to select only one primary factor which would fundamentally misrepresenting the multifaceted nature of aviation safety incidents.

Also the multi-label framework enables robust detection and mitigation of model hallucinations. Hallucination is an inevitable limitation of large language models \cite{xu2024hallucination} and as a result, it is crucial to design systems that can detect and penalize such occurrences rather than ignore them. Binary approaches fail to address the critical issue of models generating invalid or fabricated HFACS codes that do not exist in the official taxonomy. The multi-label classification framework allows us to design reward functions that explicitly penalize the generation of hallucinated codes, thereby improving model reliability and preventing the propagation of erroneous safety classifications that could mislead investigators.

\subsection{Dataset and Synthetic Data Generation}

We utilize the GAHFACS dataset \cite{liu2025accident}, which was constructed from the National Transportation Safety Board (NTSB) database covering general aviation accidents from 2008 to 2024. The GAHFACS dataset exhibits significant class imbalance, with some HFACS categories being severely underrepresented. Traditional approaches to handling imbalanced datasets in machine learning rely on sampling techniques such as SMOTE \cite{chawla2002smote}, SMOTE-ENN \cite{lamari2020smote}, Borderline-SMOTE \cite{han2005borderline}, SVM-SMOTE \cite{demidova2017svm}, SMOTE-Tomek \cite{hairani2023improvement}, and random oversampling \cite{hayaty2020random}. These methods work effectively for numerical data where synthetic samples can be generated through mathematical operations like interpolation and averaging between existing data points.

However, these conventional oversampling techniques are fundamentally incompatible with natural language processing tasks. Unlike numerical features, textual data such as accident narratives cannot be meaningfully averaged or interpolated to create synthetic examples. The discrete and symbolic nature of language makes it impossible to apply geometric transformations that form the basis of traditional oversampling methods.

\begin{figure*}[t]
\centering
\includegraphics[width=0.7\textwidth]{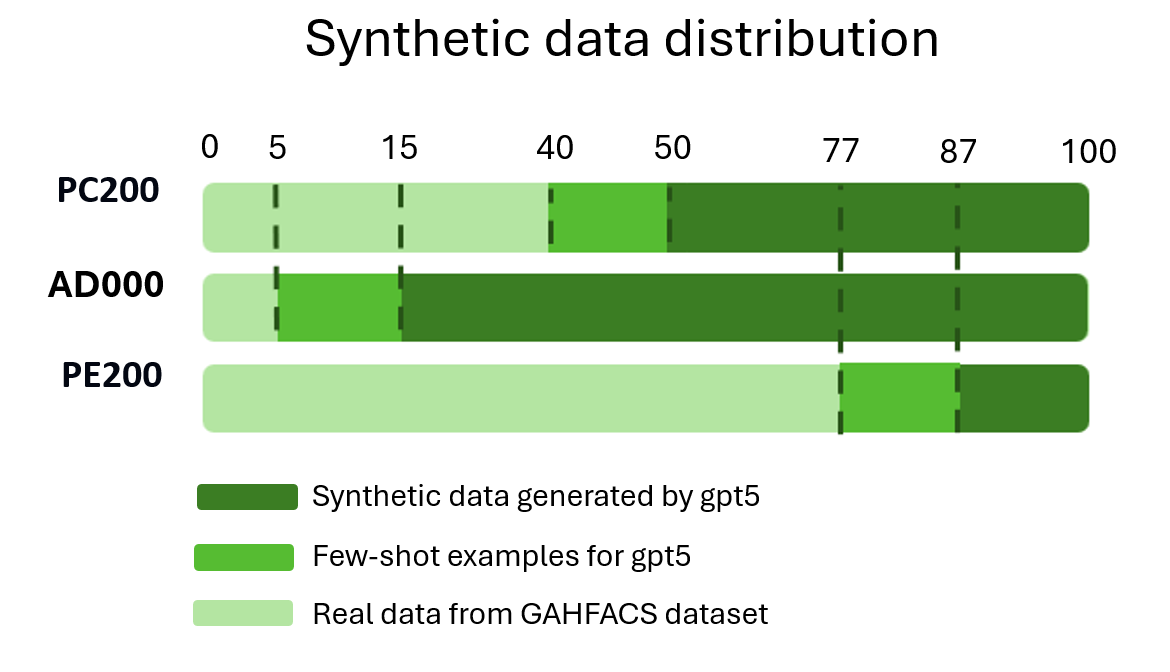}
\caption{Synthetic data distribution showing the composition of real GAHFACS data, few-shot examples for GPT-5, and synthetic data generated for underrepresented HFACS categories (PC200, AD000, PE200). The darker green represents synthetic data generated by the LLM, medium green shows few-shot examples, and light green indicates real data from the GAHFACS dataset.}
\label{fig:synthetic_data}
\end{figure*}

To address this challenge, we developed a synthetic data generation pipeline using GPT-5 that leverages few-shot learning principles. The synthetic data generation process targets underrepresented classes, particularly AD000, PC200, and PE200. As illustrated in Figure \ref{fig:synthetic_data}, our approach combines real GAHFACS data with synthetic examples generated by the LLM and few-shot examples to achieve balanced training sets. 

Few-shot examples refer to a small number of high-quality real accident narratives from each underrepresented class that serve as templates to guide the LLM in generating realistic synthetic data. These examples demonstrate the typical characteristics and language patterns associated with each HFACS category. This approach enables the LLM to produce synthetic narratives that maintain domain authenticity and at the same time expanding the available training data for rare classes.

The training set is constructed with exactly 100 samples per HFACS category (total of 1000 training samples) which creates a balanced dataset and enables the model to learn distinctive patterns within each category. When insufficient real data exists for a category, synthetic samples are generated to reach the target count. This balanced approach ensures that the model receives equal exposure to all HFACS categories during training that would prevent bias toward more frequent categories and enabling comprehensive pattern recognition across all safety factors.

The test set comprises 100 samples with an intentionally imbalanced distribution that includes at least 10\% representation from underrepresented classes. This test set size aligns with established practices in language model evaluation where similar studies have employed comparable sample sizes for assessment \cite{devlin2019bert,brown2020language,petruzzellis2024benchmarking}, and is also chosen due to practical constraints of computational cost and evaluation time for comprehensive model assessment across multiple metrics and comparison baselines. The imbalanced test set composition is deliberately designed to simulate real-world deployment conditions, as aviation safety datasets exhibit natural class imbalance due to the varying frequency of different human factors in actual accidents, thereby testing the model's ability to generalize from balanced training data to realistic operational scenarios.

\subsection{Group Relative Policy Optimization}

Group Relative Policy Optimization (GRPO) \cite{shao2024deepseekmath} is a reinforcement learning algorithm that extends Proximal Policy Optimization (PPO) \cite{schulman2017proximal} while eliminating the need for a separate value function . Instead of training a critic model, GRPO estimates advantages using relative comparisons within groups of generated responses.

\subsubsection{GRPO Formulation}

For each input prompt $q$, GRPO samples a group of $G$ outputs $\{o_1, o_2, ..., o_G\}$ from the policy model $\pi_\theta$. The algorithm optimizes the following objective:

\begin{align}
J_{GRPO}(\theta) &= \mathbb{E}_{q \sim P(Q), \{o_i\}_{i=1}^G \sim \pi_{\theta_{old}}(O|q)} \left[ \frac{1}{G} \sum_{i=1}^G \frac{1}{|o_i|} \sum_{t=1}^{|o_i|} L_i^t \right]
\end{align}

where $L_i^t$ is defined as:

\begin{align}
L_i^t &= \min\left(r_t \hat{A}_{i,t}, \text{clip}(r_t, 1-\epsilon, 1+\epsilon) \hat{A}_{i,t}\right)
\end{align}
where $r_t = \frac{\pi_\theta(o_{i,t}|q, o_{i,<t})}{\pi_{\theta_{old}}(o_{i,t}|q, o_{i,<t})}$ is the probability ratio.

The advantage $\hat{A}_{i,t}$ is computed using group-relative normalization:

\begin{align}
\hat{A}_{i,t} &= \frac{r_i - \text{mean}(\mathbf{r})}{\text{std}(\mathbf{r})}
\end{align}

where $\mathbf{r} = \{r_1, r_2, ..., r_G\}$ are the rewards for all outputs in the group, and $r_i$ is the total reward for output $o_i$.

\subsubsection{Advantage Estimation}

\begin{figure*}[t]
\centering
\includegraphics[width=0.9\textwidth]{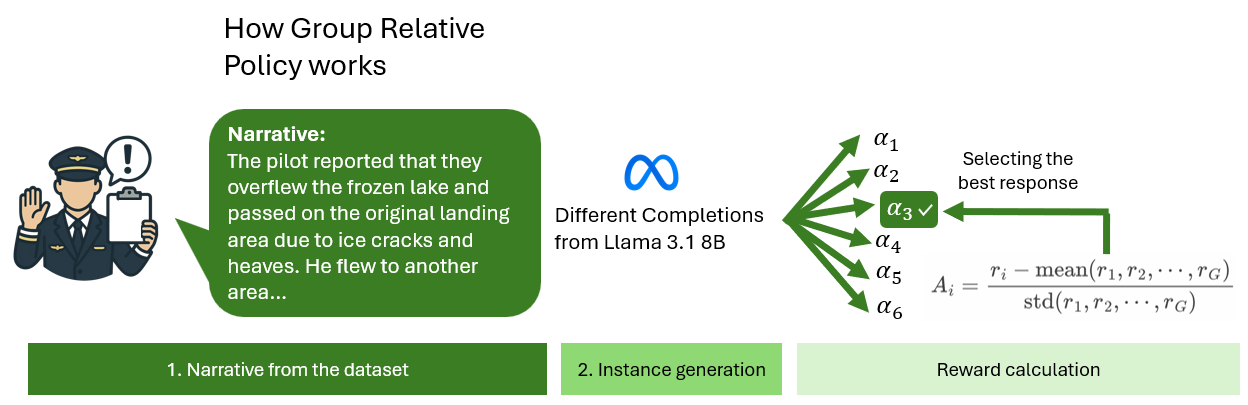}
\caption{Reward Calculation: Given an accident narrative, the model generates multiple completions ($\alpha_1$ through $\alpha_6$). Each completion receives a reward $r_i$, and the advantage $A_i$ is calculated using group-relative normalization. The best response (highlighted with checkmark) receives the highest advantage and guides policy optimization toward higher-quality outputs.}
\label{fig:grpo_mechanism}
\end{figure*}

Figure \ref{fig:grpo_mechanism} illustrates how GRPO computes advantages using group-relative comparisons. Given a narrative input, the Llama 3.1 8B generates multiple completions and each of them receives a reward score. The advantage for each completion is calculated relative to the group mean and standard deviation to enable the algorithm to identify and reinforce higher-quality responses while penalizing lower-quality ones.

\subsection{Reward Function Design}

Our reward function combines multiple components to evaluate response quality comprehensively:

\begin{align}
r_{total} &= r_{correctness} + r_{partial} + r_{format} + r_{validity} + r_{gpt5judge}
\end{align}

\noindent{a) Correctness Reward:} The primary reward component assigns +2.0 for exact matches between predicted and ground truth HFACS codes, and 0 otherwise:

\begin{align}
r_{correctness} = \begin{cases}
2.0 & \text{if exact match} \\
0.0 & \text{otherwise}
\end{cases}
\end{align}

\noindent{b) Partial Match Reward:} This component provides scaled rewards for partially correct predictions:

\begin{align}
r_{partial} = \begin{cases}
0.1 + 0.9 \times \frac{|\text{predicted} \cap \text{true}|}{|\text{true}|} & \text{if partial overlap} \\
0.0 & \text{if no overlap}
\end{cases}
\end{align}

\noindent{c) Format Reward:} This reward ensures responses follow the required structure with reasoning tags and proper HFACS code placement:

\begin{align}
r_{format} = \begin{cases}
0.25 & \text{if proper format} \\
0.0 & \text{otherwise}
\end{cases}
\end{align}

\noindent{d) Validity Reward:} This penalty targets outputs containing invalid HFACS codes:

\begin{align}
r_{validity} &= -0.25 \times |\text{invalid codes generated}|
\end{align}

\noindent{e) LLM Reasoning Quality Reward:} This component employs GPT-5-nano to evaluate reasoning quality on a scale from 0.0 to 0.5. The string inside of the reasoning tag from the LLM response is evaluated to assess logical coherence and relevance to the accident narrative.

\begin{align}
r_{gpt5} = \begin{cases}
0.5 & \text{if good reasoning} \\
0.25 & \text{if okay reasoning} \\
0 & \text{if bad reasoning}
\end{cases}
\end{align}

\subsection{Training Pipeline}
The code implements a data--reward--optimize loop. Data are loaded with a custom \texttt{GAHFACSDataLoader} that reads from the dataset, builds chat-style prompts, and performs a constrained split. Concretely, the loader excludes synthetic rows (those whose \texttt{ev\_id} begins with \texttt{SYNTH-}) from the test set, then forms a test set of 100 uniformly sampled \emph{real} items plus \%10 additional \emph{real} data from underrepresented classes \texttt{AD000}, \texttt{PC200}, and \texttt{PE200}. The training set is balanced to exactly 100 instances per HFACS code. When a deficit is detected, the loader generates the synthetic dataset using few-shot style guidance drawn from up to ten real narratives of the target class and assigns \texttt{ev\_id}s of the form \texttt{SYNTH-\{code\}-k}; an optional \texttt{--save-synthetic-data} flag writes these to Excel for audit. Each training example is a chat with a system prompt enumerating the HFACS taxonomy and formatting rules, and a user message of the form “\emph{Analyze this accident narrative:} \texttt{\$\{narr\_accf\}}”; labels are stored as a single space-separated string of codes.

Optimization uses \texttt{TRL}'s \texttt{GRPOTrainer} with the reward functions implemented in the script. During training we log, per sample, the prompt, model output, parsed codes, each reward component, and the total; logs are flushed to JSONL (full fidelity) and, when available, Parquet and Excel with long-cell sanitization. A lightweight telemetry thread collects CPU/GPU utilization. Evaluation uses \texttt{vLLM} \cite{kwon2023efficient} with temperature \(1.0\) (as of the writing of this paper, GPT-5 only supports the value 1 for tempreture, so we use this number for all of the other LLMs for consistency), \(top\_p{=}0.95\), and up to \(1024\) generation tokens; predictions are parsed from the text after the \texttt{</reasoning>} string and metrics include exact match, partial credit, and macro/micro F\(_1\).

\begin{algorithm}[!t]
  \caption{GRPO Training for HFACS Classification}
  \label{alg:grpo-hfacs}
  \small
  \begin{algorithmic}[1]
  \REQUIRE Balanced train prompts $\{(q,y)\}$ from GAHFACS with synthetic items for rare classes; group size $G{=}6$; max steps $T{=}1000$; base policy $\pi_{\theta_{\mathrm{init}}}$; tokenizer; reference policy for KL.
  
  \STATE \textbf{Initialize model:} Load \texttt{meta-Llama-3.1-8B} in 4-bit; attach LoRA with rank $32$ on \texttt{q\_proj}, \texttt{k\_proj}, \texttt{v\_proj}, \texttt{o\_proj}, \texttt{gate\_proj}, \texttt{up\_proj}, \texttt{down\_proj}; set \texttt{use\_gradient\_checkpointing="unsloth"}; set max prompt $=1024$ and max completion $=1024$ tokens.
  
  \STATE \textbf{Set GRPO config:} learning rate $5{\times}10^{-6}$, warmup ratio $0.1$, weight decay $0.1$, optimizer \texttt{paged\_adamw\_8bit}, per-device batch size $1$, gradient clipping $0.1$, save every $250$ steps, output dir \texttt{outputs}.
  
  \FOR{$t = 1$ to $T$}
    \STATE Sample a minibatch of prompts $q$ and set $\pi_{\theta_{\mathrm{old}}} \leftarrow \pi_\theta$.
    \STATE For each $q$, sample $G$ completions $o_1,\ldots,o_G \sim \pi_{\theta_{\mathrm{old}}}(\cdot \mid q)$ constrained to the format \texttt{<reasoning>}…\texttt{</reasoning>} followed by space-separated HFACS codes.
      \STATE For each $o_i$, compute rewards on the post-\texttt{</reasoning>} segment:
  \STATE \hspace{1em} $r_{\text{correct}} = 2.0$ if predicted set equals $y$; else $0$.
  \STATE \hspace{1em} $r_{\text{partial}} = 0.1 + 0.9 \cdot \frac{\lvert \text{pred} \cap y \rvert}{\lvert y \rvert}$ if $0<\lvert \text{pred}\cap y\rvert<\lvert y\rvert$; else $0$.
  \STATE \hspace{1em} $r_{\text{format}} = 0.25$ if \texttt{<reasoning>}…\texttt{</reasoning>} exists and codes occur after \texttt{</reasoning>}; else $0$.
  \STATE \hspace{1em} $r_{\text{valid}} = -0.25 \times \lvert\{\text{invalid HFACS tokens}\}\rvert$.
  \STATE \hspace{1em} Extract reasoning text from \texttt{<reasoning>}…\texttt{</reasoning>} tags.
  \STATE \hspace{1em} $r_{\text{gpt5}} = \text{GPT-5-nano}(\text{reasoning\_text}, \text{accident\_narrative})$ where output $\in \{0, 0.25, 0.5\}$ based on reasoning quality (bad, okay, good).
  \STATE \hspace{1em} $r_i = r_{\text{correct}} + r_{\text{partial}} + r_{\text{format}} + r_{\text{valid}} + r_{\text{gpt5}}$.
    \STATE Compute group-relative advantages $\hat{A}_{i,t} = \big(r_i - \mathrm{mean}(\mathbf{r})\big)/\mathrm{std}(\mathbf{r})$ for tokens of $o_i$.
    \STATE Update $\pi_\theta$ by maximizing the clipped PPO ratio with tokenwise advantages and a KL penalty to the reference policy (GRPO update).
  \ENDFOR
  
  \STATE Periodically checkpoint LoRA weights and persist JSON/CSV logs of prompts, generations, and reward components for auditability.
  \end{algorithmic}
  \end{algorithm}  

\subsection{Implementation Details}
We load \texttt{meta-Llama-3.1-8B-In\-struct} with FastLanguageModel.\-from pre\-train\-ed (\texttt{load\_in\_4bit=True}, \texttt{fast\_inference=True}, \texttt{max\_lora\_rank=32}, \texttt{gpu\_memory\_utilization=0.8}), and attach LoRA adapters of rank \(32\) to \texttt{q\_proj}, \texttt{k\_proj}, \texttt{v\_proj}, \texttt{o\_proj}, \texttt{gate\_proj}, \texttt{up\_proj}, and \texttt{down\_proj}, with \texttt{use\_gradient\_check\-point\-ing="unsloth"} and a \texttt{max\_seq\_length} of \(2048\) split as \(1024{+}1024\) for prompt and completion. Training uses \texttt{GRPOTrainer} with \texttt{num\_generations=6}, learning rate \(5\!\times\!10^{-6}\), cosine decay, warmup ratio \(0.1\), weight decay \(0.1\), \texttt{paged\_adamw\_8bit}, per-device batch size \(1\), accumulation \(1\), gradient clipping \(0.1\), 1000 steps of training, and checkpoints every \(250\) steps to \texttt{outputs}. Prompts are rendered with \texttt{tokenizer.apply\_chat\_tem\-plate(\ldots,\ add\_generation\_prompt=True)}. Evaluation uses \texttt{vLLM} \cite{kwon2023efficient} sampling (\(T{=}1.0\), \(p{=}0.95\), \(1024\) tokens). Reward logs contain both raw outputs and parsed fields, and are persisted in multiple formats; a background monitor records CPU/GPU utilization during training.

\section{Results and Discussion}

\subsection{Performance Improvements with GRPO}

Our experiments demonstrate that GRPO optimization noticeably enhances HFACS classification performance across all evaluation metrics. Code and data are available at: https://github.com/INQUIRELAB/llm-rl-grpo-aviation-analysis. Given the multi-label nature of the classification problem, we assess performance using two metrics: Exact Match Accuracy and Partial Match Accuracy. Exact match accuracy measures the proportion of predictions where the model's output HFACS codes exactly match the ground truth set. It requires both precision and recall of 100\% for each sample. Partial match accuracy measures the proportion of predictions where at least one predicted HFACS code matches a code in the ground truth set which provides a more lenient assessment of the model's ability to identify relevant safety factors.

\begin{figure}[t]
    \centering
    \includegraphics[width=\columnwidth]{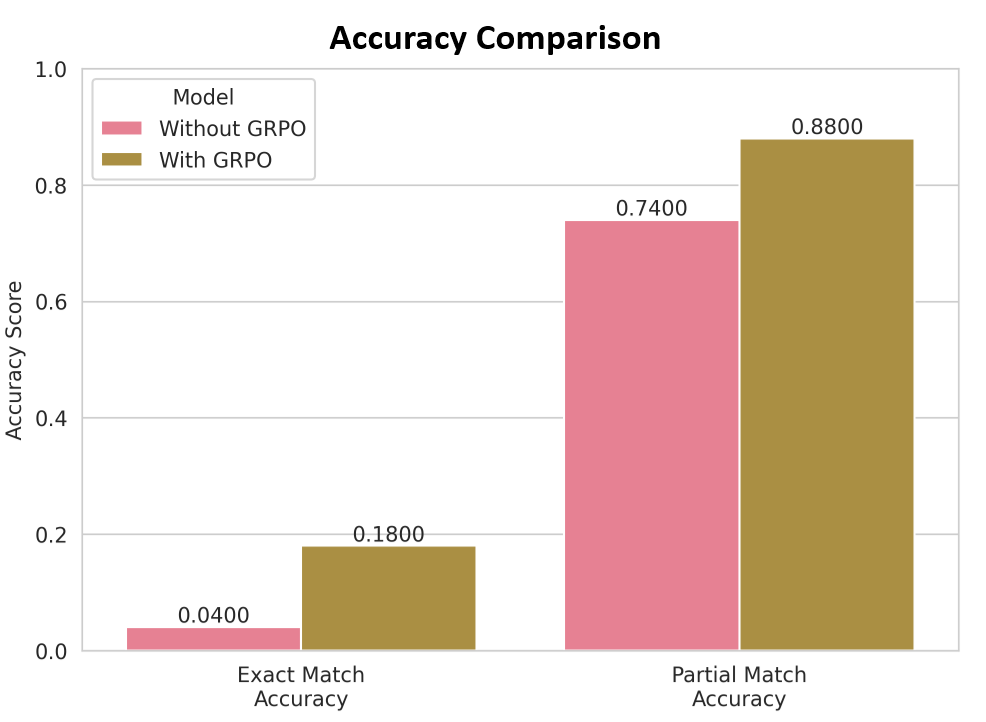}
    \caption{Comparison of exact match and partial match accuracy between baseline and GRPO-optimized models. GRPO training yields substantial improvements in both metrics.}
    \label{fig:accuracy_comparison}
\end{figure}

Figure~\ref{fig:accuracy_comparison} illustrates the most noticeable improvements: exact match accuracy increased from 0.04 to 0.18. A 350\% improvement, while partial match accuracy improved from 0.74 to 0.88.

\begin{figure}[t]
    \centering
    \includegraphics[width=\columnwidth]{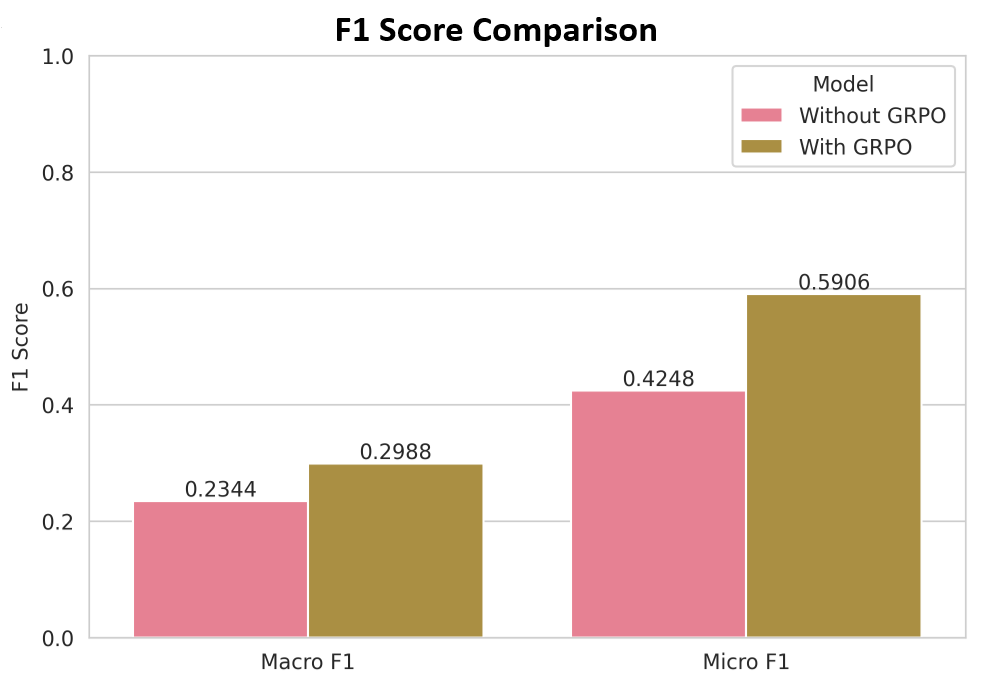}
    \caption{F1 score improvements with GRPO optimization. Both macro and micro F1 scores show substantial gains, with micro F1 demonstrating particularly strong improvement.}
    \label{fig:f1_comparison}
\end{figure}

The improvements extend beyond raw accuracy to more nuanced classification metrics. As shown in Figure~\ref{fig:f1_comparison}, the macro F1 score increased from 0.2344 to 0.2988, while the micro F1 score improved from 0.4248 to 0.5906. The distinction between these metrics is crucial for understanding model performance: Macro F1 calculates the F1 score for each HFACS category independently and then averages them that gives equal weight to all categories regardless of their frequency. Micro F1 aggregates all true positives, false positives, and false negatives across all categories before calculating the F1 score, thus giving more weight to frequent categories. In the context of aviation safety, macro metrics ensure that rare but critical failure modes are not overlooked, while micro metrics reflect overall system performance on typical incidents.

\subsection{Detailed Performance Analysis}

\begin{figure}[t]
    \centering
    \includegraphics[width=\columnwidth]{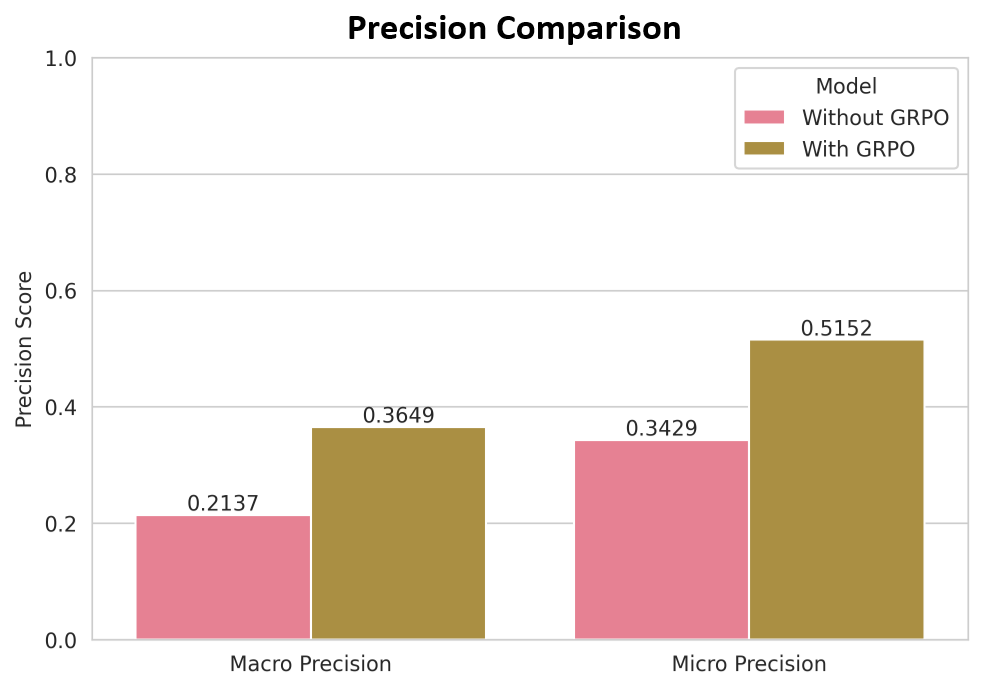}
    \caption{Precision improvements with GRPO showing substantial gains across macro and micro calculations.}
    \label{fig:precision_comparison}
\end{figure}

\begin{figure}[t]
    \centering
    \includegraphics[width=\columnwidth]{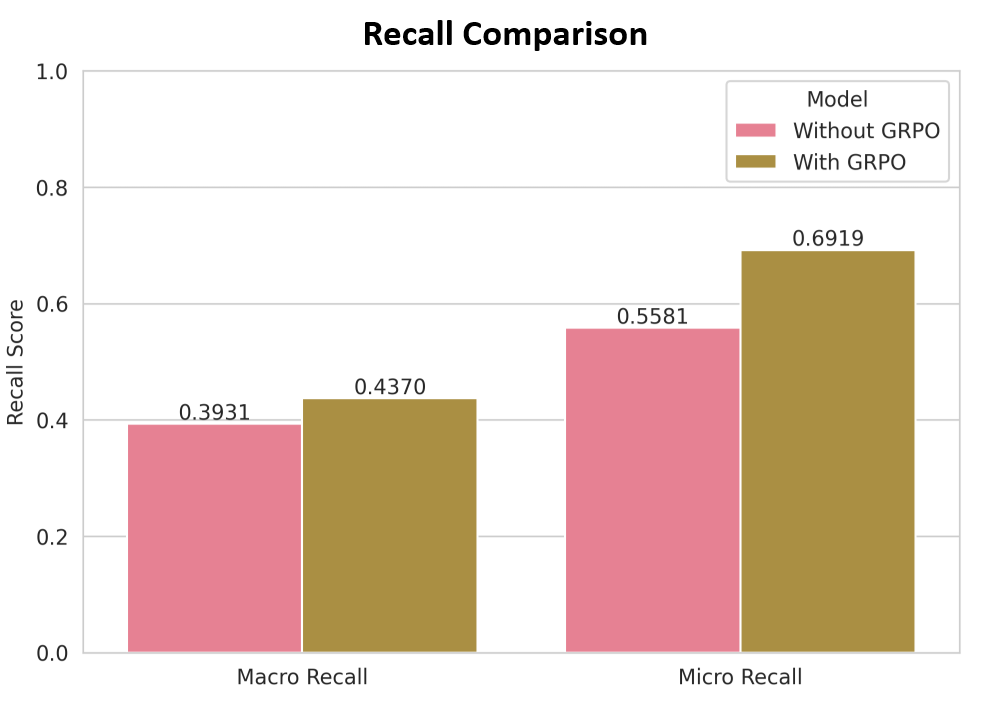}
    \caption{Recall improvements with GRPO demonstrating consistent gains in both macro and micro metrics.}
    \label{fig:recall_comparison}
\end{figure}

Figures~\ref{fig:precision_comparison} and \ref{fig:recall_comparison} reveal complementary improvements in precision and recall. Macro precision increased from 0.2137 to 0.3649 (71\% improvement), while micro precision rose from 0.3429 to 0.5152. Similarly, recall metrics showed consistent gains, with macro recall improving from 0.3931 to 0.4370 and micro recall from 0.5581 to 0.6919. The balanced improvements across precision and recall basically shows GRPO optimization does not simply trade one metric for another but genuinely enhances the model's ability to identify relevant HFACS codes while reducing false positives.

\subsection{Comparison with other Large Language Models}

Table~\ref{tab:model_comparison} presents a comprehensive comparison of our GRPO-optimized model against state-of-the-art language models. Our GRPO-optimized model achieves the highest exact match accuracy (0.1900) across all evaluated models, outperforming Gemini-2.5-flash (0.1200), GPT-5-mini (0.0700), and significantly surpassing larger open source models including the 20B parameter GPT-OSS (0.0600). The model also ties with GPT-5-mini for the highest partial match accuracy (0.8800) while surpassing Gemini-2.5-flash (0.8300). Additionally, our model achieves the highest micro F1 score (0.5906) and micro precision (0.5152) across all evaluated models, with macro precision (0.3649) closely approaching Gemini-2.5-flash's performance (0.3691). While these exact match accuracy percentages may appear modest, such performance levels are characteristic of complex reasoning tasks for current LLMs. The ARC-AGI-2 benchmark \cite{chollet2025arc} demonstrates this pattern, where state-of-the-art models like GPT-5 Mini (High) achieve only 4.4\% accuracy and Gemini 2.5 Pro variants score around 2.1-2.5\%, while humans can achieve near-perfect performance on the same tasks. This suggests that HFACS classification, like ARC-AGI-2, has the potential to serve as a valuable real-world benchmarking method for evaluating the reasoning capabilities of different LLMs, as it presents challenges that are manageable for human experts but difficult for current AI systems.

While Google DeepMind and OpenAI have not disclosed the parameter counts for Gemini-2.5-flash and GPT-5-mini respectively, we can estimate their scale using the LMSYS benchmark \cite{zheng2023lmsys}. As of this writing, Gemini-2.5-flash achieves a score of 1406, which closely matches the open-source Qwen3-235b-a22b-thinking-2507 model (1403 points, 235 billion parameters), while GPT-5-mini scores 1390 points, both substantially outperforming the baseline Llama 3.1 8B model (approximately 1215 points on the same benchmark). We were able to demonstrate that our much smaller, domain-optimized model can outperform these significantly larger general-purpose models on specialized tasks through targeted reinforcement learning, while also enabling deployment on resource-constrained edge devices where larger models would be impractical.

\begin{table*}[t]
    \centering
    \caption{Performance comparison across different language models for HFACS classification}
    \label{tab:model_comparison}
    \footnotesize
    \begin{tabular}{lccccccc}
        \hline
        \textbf{Model} & \textbf{Params} & \textbf{Exact} & \textbf{Partial} & \textbf{Macro F1} & \textbf{Macro Prec.} & \textbf{Micro F1} & \textbf{Micro Prec.} \\
        \hline
        Llama-3.1 (Baseline) & 8B & 0.0400 & 0.7400 & 0.2344 & 0.2137 & 0.4248 & 0.3429 \\
        Llama-3.1 (GRPO) & 8B & \textbf{0.1800} & \textbf{0.8800} & 0.2988 & 0.3649 & \textbf{0.5906} & \textbf{0.5152} \\
        \hline
        GPT-5-mini & - & 0.0700 & \textbf{0.8800} & \textbf{0.3733} & 0.3174 & 0.5070 & 0.3625 \\
        Gemini-2.5-flash & - & 0.1200 & 0.8300 & 0.3730 & \textbf{0.3691} & 0.5459 & 0.4585 \\
        \hline
        GPT-OSS & 20B & 0.0600 & 0.8400 & 0.3429 & 0.3422 & 0.4737 & 0.3500 \\
        Gemma3 & 12B & 0.0700 & 0.6700 & 0.2686 & 0.2913 & 0.4038 & 0.3443 \\
        DeepSeek-R1 & 14B & 0.0600 & 0.6800 & 0.2700 & 0.2933 & 0.4320 & 0.3708 \\
        Gemma3n-e4b & 6.8B & 0.0200 & 0.4300 & 0.2406 & 0.2656 & 0.2507 & 0.2275 \\
        Qwen3 & 8B & 0.0700 & 0.7000 & 0.2802 & 0.3162 & 0.4172 & 0.3420 \\
        \hline
    \end{tabular}
\end{table*}

\subsection{Training Dynamics and Reward Evolution}

\begin{figure*}[t]
    \centering
    \subfigure[Total reward evolution]{
        \includegraphics[width=0.45\textwidth]{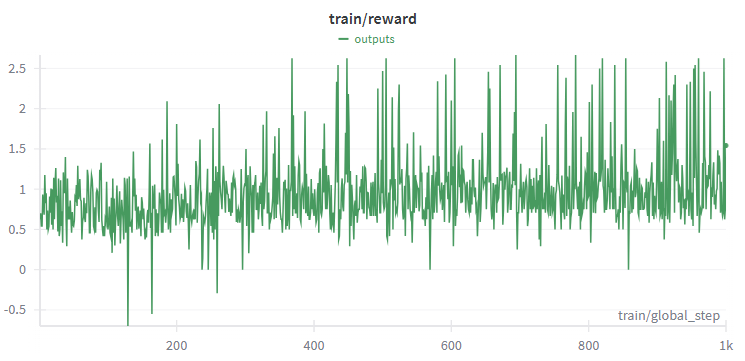}
    }
    \hfill
    \subfigure[Correctness reward progression]{
        \includegraphics[width=0.45\textwidth]{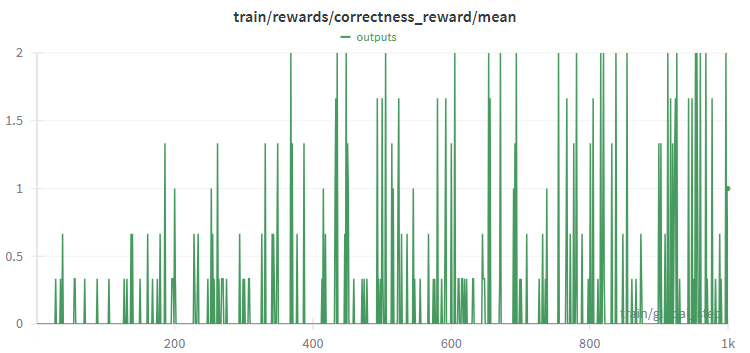}
    }
    \caption{Training dynamics showing reward evolution over 1000 steps. The consistent improvement and stabilization around step 420 indicate effective optimization.}
    \label{fig:reward_evolution}
\end{figure*}

The training progression reveals important insights into how GRPO optimization improves model behavior. Figure~\ref{fig:reward_evolution} displays the evolution of total reward and correctness reward during training. The total reward exhibits a consistent upward trend that over time stabilizes around step 420, while the correctness reward shows gradual improvement with notable spikes corresponding to breakthrough moments where the model discovers better classification strategies to improve its exact match accuracy.

\begin{figure}[t]
    \centering
    \includegraphics[width=\columnwidth]{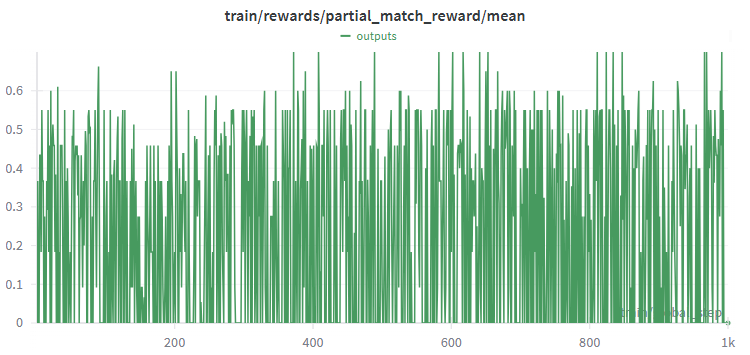}
    \caption{Evolution of partial match reward during training that has consistent performance in identifying relevant HFACS categories.}
    \label{fig:partial_match_reward}
\end{figure}

\begin{figure}[t]
    \centering
    \includegraphics[width=\columnwidth]{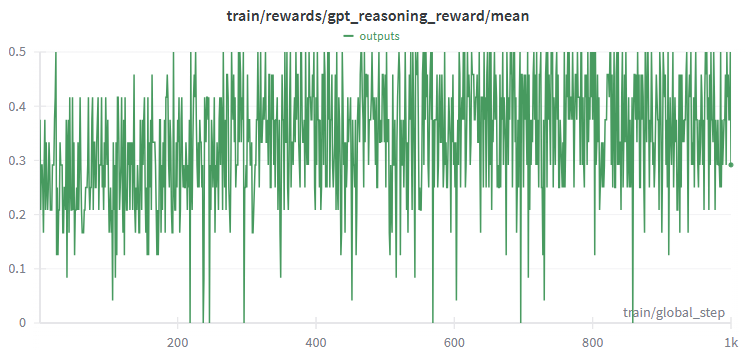}
    \caption{GPT-5 reasoning quality reward evolution.}
    \label{fig:reasoning_reward}
\end{figure}

Figure~\ref{fig:partial_match_reward} shows partial match reward evolution, which maintains relatively high values throughout training. This suggests that the model quickly learns to identify relevant HFACS categories even when not achieving perfect matches. The GPT-5 reasoning quality reward (Figure~\ref{fig:reasoning_reward}) shows variable but generally improving trends and it will show the model's explanatory capabilities develop alongside its classification accuracy.

\subsection{Computational Efficiency and Edge Deployment}

\begin{figure}[t]
    \centering
    \includegraphics[width=\columnwidth]{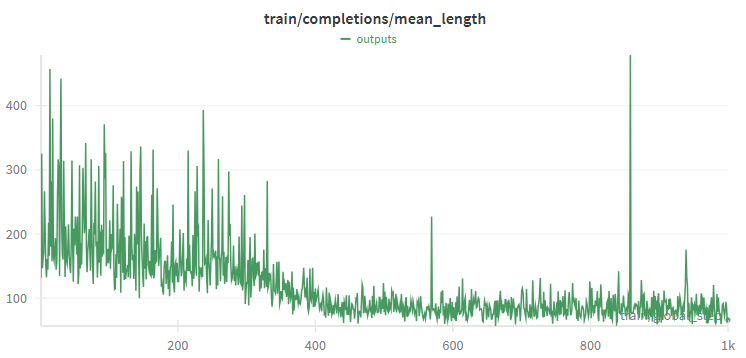}
    \caption{Evolution of mean completion length during training. The model learns to generate more concise responses while improving classification accuracy.}
    \label{fig:completion_length}
\end{figure}

The completion length analysis (Figure~\ref{fig:completion_length}) reveals an interesting training dynamic: initial responses are verbose (averaging 350 tokens), but the model learns to generate more concise outputs (stabilizing around 100 tokens) while maintaining or improving accuracy. This efficiency gain has practical implications for deployment. The final outcome is that both computational costs and response latency are reduced substantially.

\begin{figure}[t]
  \centering
  \includegraphics[width=\columnwidth]{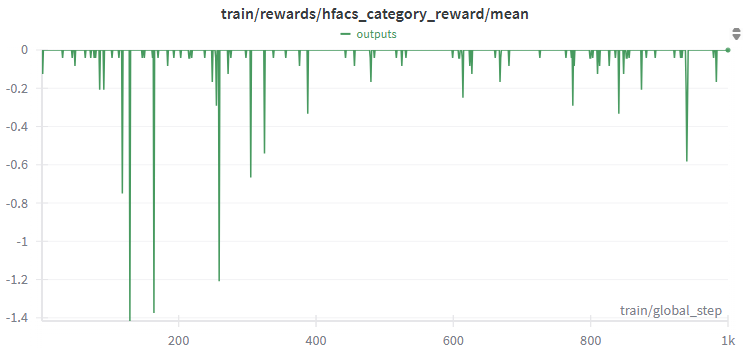}
  \caption{The HFACS category reward function penalizes the model for hallucinating HFACS categories. The model hallucinates less after 400 steps.}
  \label{fig:category_reward}
\end{figure}

The HFACS category reward evolution (Figure \ref{fig:category_reward}) reveals a clear learning pattern in the model's tendency to hallucinate invalid codes. During the initial 400 training steps, the model receives substantial penalties (reaching -1.4) as it frequently generates non-existent HFACS categories. However, after step 400, the penalties significantly decrease and stabilize around -0.2 to -0.4, indicating that the model learns to reduce hallucination behavior and generate more valid HFACS codes. This improvement coincides with the overall reward stabilization observed in other metrics, suggesting that reducing hallucination is a critical component of the model's learning process during GRPO optimization.

\begin{figure}[t]
  \centering
  \includegraphics[width=\columnwidth]{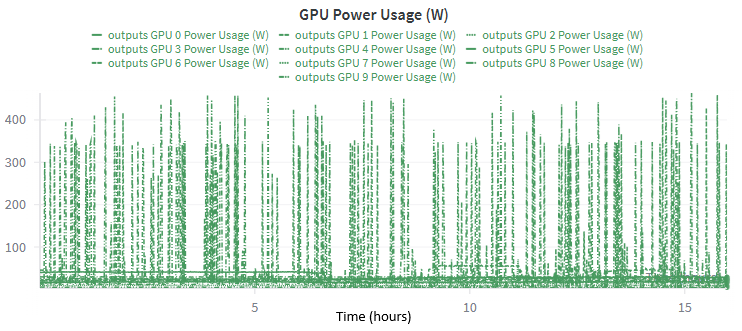}
  \caption{GPU power consumption during training. This figure shows stable resource utilization with periodic spikes for checkpointing.}
  \label{fig:gpu_usage}
\end{figure}

GPU power usage (Figure~\ref{fig:gpu_usage}) remained relatively stable throughout training, with periodic spikes corresponding to checkpoint saves and evaluation runs. The consistent power consumption around 300-400W shows the computational feasibility of GRPO training on standard hardware. Importantly, the trained GRPO model was also tested on NVIDIA Jetson Orin Nano in MAXN Super mode with just 25W power consumption which proves this approach is suitable for resource-constrained environments.

\begin{figure}[t]
    \centering
    \includegraphics[width=\columnwidth]{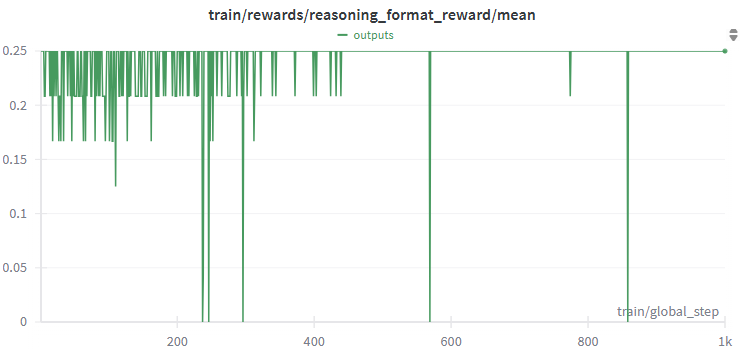}
    \caption{Format reward showing the model becomes better at putting the reasoning string in the correct tag. This is crucial for the model to become a reasoning model.}
    \label{fig:format_reward}
\end{figure}

The format reward evolution (Figure~\ref{fig:format_reward}) shows that the LLM gradually learns how to use the reasoning tags after 420 which make the model to be a reasoning model overtime. 

\section{Conclusion}
In this research, we have demonstrated the novel application of Group Relative Policy Optimization (GRPO) to enhance the automated classification of aviation accidents using the Human Factors Analysis and Classification System (HFACS). Our approach addresses the critical need for computationally efficient and accurate safety analysis by fine-tuning a comparatively small 8B parameter language model, Llama-3.1, which ultimately outperformed larger, general-purpose state-of-the-art models.

A cornerstone of our methodology is a comprehensive, multi-component reward system designed to capture the specific requirements of HFACS classification. With integrating rewards for exact and partial matches, format compliance, and reasoning quality (while simultaneously penalizing the hallucination of invalid codes), we then guided the model toward generating methodologically rigorous and reliable outputs. Furthermore, we effectively mitigated the pervasive issue of class imbalance in aviation safety datasets by developing a synthetic data generation pipeline that ensures the model was adequately trained on rare yet critical accident categories.

Our experimental results show substantial performance gains. The GRPO-optimized model achieved a 350\% improvement in exact match accuracy and a significant increase in partial match accuracy from 0.74 to 0.88 compared to its baseline. This specialized model not only surpassed its own baseline but also demonstrated superior performance in key metrics against significantly larger models like Gemini-2.5-flash and GPT-5-mini, highlighting the power of targeted reinforcement learning for specialized domains. Analysis of the training dynamics revealed that the model learned to produce more concise and accurate classifications over time.

The primary contributions of this work are fourfold: (1) the first successful application of GRPO to an aviation safety classification task; (2) the demonstration that a smaller, domain-optimized model can outperform larger general-purpose models; (3) the development of a sophisticated, multi-faceted reward function tailored for HFACS analysis; and (4) the introduction of exact match accuracy for multi-label HFACS classification as a challenging new benchmark for LLM reasoning capabilities.

Ultimately, this research validates a pathway toward developing more accurate and efficient automated safety analysis tools that are practical for real-world deployment, including on resource-constrained edge devices. Future work could extend this methodology to other high-stakes domains, explore more advanced reward-modeling techniques, and being able to interpret and trace the model's circuits for analysing the outputs and enabling interpretability of model reasoning pathways.


\section{Data Statement}
The GAHFACS dataset used in this study was obtained by contacting the authors of \cite{liu2025accident} who constructed it from publicly available National Transportation Safety Board (NTSB) accident reports covering general aviation incidents from 2008 to 2024. The processed dataset and synthetic data generation code developed for this study are available at: \url{https://github.com/INQUIRELAB/llm-rl-grpo-aviation-analysis}. Raw NTSB data can be accessed through the official NTSB database. All synthetic data used for class balancing is clearly labeled and separated from real accident data to ensure transparency and reproducibility.

\section{CRediT Author Statement}
\textbf{Arash Ahmadi:} Conceptualization, Methodology, Software, Validation, Formal analysis, Investigation, Data curation, Writing – original draft, Writing – review \& editing, Visualization, Project administration. \textbf{Sarah S. Sharif:} Conceptualization, Resources, Writing – review \& editing, Supervision \textbf{Yaser M. Banad:} Conceptualization, Resources, Writing – review \& editing, Supervision

\section{Declaration of Competing Interests}
The authors declare that they have no known competing financial interests or personal relationships that could have appeared to influence the work reported in this paper.

\section{Declaration of Generative AI and AI-assisted Technologies in the Writing Process}
During the preparation of this work, the authors used GPT-5 for multiple purposes within the research methodology. GPT-5 was employed for synthetic data generation to address class imbalance in aviation safety datasets, where it generated realistic accident narratives for underrepresented HFACS categories including AD000, PC200, and PE200 using few-shot learning approaches with real accident examples as templates. Additionally, GPT-5-nano served as a reasoning quality evaluator within our multi-component reward system which assesses the logical coherence and relevance of model-generated explanations during the reinforcement learning training process. This LLM judge component provided reward scores ranging from 0.0 to 0.5 based on the quality of reasoning contained within the model's chain-of-thought explanations. For model evaluation and benchmarking, we utilized several large language models including GPT-5-mini, Gemini-2.5-flash, GPT-OSS, Gemma3, DeepSeek-R1, Gemma3n-e4b, and Qwen3 to establish performance baselines and demonstrate the effectiveness of our GRPO-optimized approach. The synthetic data generation process and LLM evaluation methodology are described comprehensively in the methodology section. The authors also used Grammarly for grammar and spell-checking assistance during manuscript preparation. After using these tools and services, the authors reviewed and edited all content as needed and take full responsibility for the content of the published article.

\section{Acknowledgements}
We want to acknowledge the National Transportation Safety Board for maintaining the comprehensive aviation accident database that made this research possible.

\section{Funding}
This research did not receive any specific grant from funding agencies in the public, commercial, or not-for-profit sectors.

\section{Glossary}
\begin{description}
\item[HFACS] Human Factors Analysis and Classification System - A systematic framework for identifying and classifying human errors in aviation accidents
\item[GRPO] Group Relative Policy Optimization - A reinforcement learning algorithm that estimates advantages using relative comparisons within groups
\item[LoRA] Low-Rank Adaptation - A parameter-efficient fine-tuning technique for large language models  
\item[GAHFACS] General Aviation Human Factors Analysis and Classification System dataset
\item[NTSB] National Transportation Safety Board - US federal agency investigating transportation accidents
\item[LLM] Large Language Model - Deep learning models trained on vast text corpora for natural language processing
\end{description}

\bibliographystyle{elsarticle-num} 
\bibliography{references}
\end{document}